\documentclass[acmsmall]{acmart}

\AtBeginDocument{%
  }

\usepackage{xspace}
\usepackage{wrapfig}
\usepackage{epsfig}
\usepackage{epstopdf}
\usepackage{listings}
\usepackage{graphicx}
\usepackage{algorithmic}
\usepackage{color}
\usepackage{comment}
\usepackage{alltt}
\usepackage{xspace}
\usepackage{epsfig,multirow, color}

\newcommand{\distancealg}{\texttt{GLB-Dist}\xspace}
\newcommand{\costalg}{\texttt{GLB-Cost}\xspace}
\newcommand{\carbonalg}{\texttt{GLB-Carbon}\xspace}

\newcommand{\ouralg}{\texttt{eGLB}\xspace}

\newcommand{\ouralgoffline}{\texttt{eGLB-Off}\xspace}

\setcopyright{none}
\renewcommand\footnotetextcopyrightpermission[1]{}
\begin{document}

\title{Towards Environmentally Equitable AI}

\author{Mohammad Hajiesmaili}
\authornote{All the authors contributed equally and are listed
in alphabetical order of last name.}
\affiliation{%
  \institution{University of Massachusetts, Amherst}
  \country{United States}
}

\author{Shaolei Ren}
\affiliation{%
  \institution{University of California, Riverside}
  \country{United States}
}

\author{Ramesh K. Sitaraman}
\affiliation{%
  \institution{University of Massachusetts, Amherst}
  \country{United States}
}

\author{Adam Wierman}
\affiliation{%
  \institution{California Institute of Technology}
  \country{United States}
  }

\renewcommand{\shortauthors}{}

\begin{abstract}
 The skyrocketing demand for artificial intelligence (AI) has created an enormous appetite for globally deployed power-hungry
servers. As a result, the environmental footprint of AI systems has come under increasing scrutiny. More crucially, the current way
that we exploit AI workloads' flexibility and
manage AI systems can lead to wildly different environmental impacts
across locations, increasingly raising environmental inequity concerns and creating unintended sociotechnical consequences. In this paper,
we advocate {environmental equity} as a priority
for the management of future AI systems,
advancing the boundaries of existing resource management for sustainable AI
 and also adding a unique dimension to AI fairness.
Concretely,
we uncover the potential of equity-aware
geographical load balancing to fairly re-distribute
the environmental cost across different regions, followed
by algorithmic challenges. We conclude by discussing a few future directions to exploit the full potential of system management approaches
to mitigate AI's environmental inequity.
\end{abstract}

\maketitle

\section{Introduction}

The growing adoption of artificial intelligence (AI) has been accelerating
across all parts of society, boosting productivity
and addressing pressing global challenges such as climate change.
Nonetheless, the technological advancement of AI
relies on computationally intensive calculations and thus
has led to a surge in resource usage 
and energy consumption. 
Even putting aside the environmental toll of server manufacturing
and supply chains,
AI systems can create a huge environmental cost to communities
and regions where they are deployed, including
air/thermal pollution 
due to fossil fuel-based electricity generation
and further stressed water resources due to AI's staggering
water footprint \cite{Carbon_SustainbleAI_CaroleWu_MLSys_2022_wu2022sustainable,Shaolei_Water_AI_Thirsty_arXiv_2023_li2023making}. To make AI  more environmentally friendly and ensure
that its overall impacts on climate change are positive, 
recent studies have pursued multi-faceted approaches, including
 efficient
training and inference \cite{ML_FrugalGPT_Cost_JamesZou_Stanford_arXiv_2023_chen2023frugalgpt},
energy-efficient GPU and accelerator designs \cite{ML_CarbonFoorptint_Google_JeffDean_Journal_2022_9810097}, carbon forecasting \cite{maji2022carboncast}, 
carbon-aware task scheduling \cite{Carbon_CarbonExplorer_HolisticGreenDataCenter_CaroleWu_BenjaminLee_ASPLOS_2023_10.1145/3575693.3575754,Google_CarbonAwareComputing_PowerSystems_2023_9770383},
green cloud infrastructures \cite{Carbon_SustainableClouds_VirtualizingEnergy_DavidIrwin_AdamWierman_SoCC_2021_10.1145/3472883.3487009},
sustainable AI policies \cite{OECD_MeasuringEnvronmentalImpacts_AI_2022_/content/paper/7babf571-en,Water_StandardSustainableAI_ISOIEC}, and more.
Additionally, data center operators have also increasingly
adopted carbon-free energy (such as solar and wind power)
and climate-conscious cooling systems,
lowering carbon footprint and direct water consumption
\cite{Google_SustainabilityReport_2023}.

Although these initiatives are encouraging, unfortunately, a worrisome outcome --- \textit{environmental inequity} --- has emerged \cite{Justice_LanguageModel_Dangers_PolicyDiscussion_TimnitGebru_FAccT_2021_10.1145/3442188.3445922}. 
That is, minimizing the total environmental cost of a globally deployed AI system across multiple regions
does not necessarily mean that each region is treated equitably. 
In fact, the environmental cost of AI is often disproportionately higher in certain disadvantaged regions than in others.
Even worse, AI's environmental inequity can be amplified by 
existing environmental equity \emph{agnostic} resource allocation, load balancing, and scheduling algorithms
and compounded by enduring socioeconomic disparities between regions.
For example, geographical load balancing (GLB) algorithms that aggressively
exploit regional differences to seek lower electricity prices 
 and/or more renewables \cite{Gao:2012:EG:2377677.2377719, murillo2024cdn}  
may schedule more workloads to water-inefficient data centers (located in, for example, water-stressed Arizona), resulting in a disproportionately high water footprint and adding further pressures to local water supplies \cite{Shaolei_Water_SpatioTemporal_GLB_TCC_2018_7420641}.

Addressing the emerging environmental inequity is becoming an integral part of responsible AI
\cite{Justice_LanguageModel_Dangers_PolicyDiscussion_TimnitGebru_FAccT_2021_10.1145/3442188.3445922}. 
It has increasingly received public attention and urgent calls
for mitigation efforts. For example, 
the AI Now Institute compares the uneven regional distribution of AI's environmental costs to ``historical practices of settler colonialism and racial capitalism'' in its 2023 Landscape report \cite{Justice_AINowInstitute_ConfrontingTechPower_2023};
the United Nations Educational, Scientific and Cultural Organization (UNESCO) recommends against the usage of AI if it creates ``disproportionate negative impacts on the environment'' \cite{Justice_Policy_Ethical_AI_Recommendation_UNESCO_2022};
California recognizes the need for
``ensuring environmental costs are equitably distributed''
in its State Report \cite{Justice_AI_EnvironmentalEquity_CaliforniaReport_2023};
and environmental justice is ranked by Meta as the most critical
factor among all environmental-related topics \cite{Facebook_SustainabilityReport_2021}.

In this paper, we advocate \emph{environmental equity} as a priority
for the management of future globally deployed AI systems. Concretely,
we explore the potential of harnessing
AI workloads' scheduling flexibility and utilizing equity-aware
GLB as a lever to fairly re-distribute
the environmental cost across regions,
ensuring that no single region disproportionately bears the environmental burden.
 Then, we present key algorithmic challenges 
to enable AI's environmental
equity without significantly degrading the other performance metrics, such as the energy cost and inference accuracy. 
Finally, we discuss future directions to unleash
the full potential of system management
for environmentally equitable AI,
including coordinated scheduling
 of AI training and inference, joint optimization of IT and non-IT resources, 
holistic control of system knobs, and building theoretical foundations.

Our proposal of environmental equity 
advances the boundaries of existing research on sustainable AI and mitigates the otherwise uneven distribution
of AI's environmental costs across different regions.
Additionally, equity and fairness are crucial considerations for 
AI.
The existing research in this space has predominantly 
tackled prediction unfairness against disadvantaged
individuals and/or groups  \cite{Fair_ML_Survey_ACM_2022_10.1145/3494672,Fair_SequentialDecision_Survey_MingyanLiu_UMich_Handbook_2021_Zhang2021}.
Thus,  environmental equity adds a unique dimension
of fairness 
and significantly complements the existing literature, 
collaboratively building equitable and responsible AI.

\section{Opportunities and Challenges for Equity-Aware GLB}

In this section, we present the potential opportunities
of leveraging equity-aware
GLB to fairly re-distribute
the environmental cost across different regions,
followed by algorithmic challenges.

\subsection{Opportunities}

The limited power grid capacity has necessitated increasing flexibility from data centers to support demand response and maintain grid stability. A notable example is the recent industry initiative to maximize load flexibility for grid-integrated data centers \cite{DataCenter_FlexibilityInitiative_EPRI_WhitePaper_2024}. Specifically, AI workloads exhibit three primary types of flexibility:  
(1) \emph{Spatial}: AI training and inference tasks can be distributed across multiple data centers with minimal impact on latency.  
(2) \emph{Temporal}: AI training tasks can be executed intermittently, provided they meet a given deadline.  
(3) \emph{Performance}: A single inference request can be processed by different AI models, each offering distinct trade-offs between accuracy and resource consumption.  
These flexibilities can be exploited to promote environmental equity while satisfying other performance objectives. To achieve this, we can leverage a variety of approaches, such as AI computing resource allocation, load balancing and job scheduling, which we
collectively refer to as system \emph{knobs}.

In practice,
the data center fleet of large companies such as Google and Microsoft often 
includes a few tens of self-managed hyperscale data centers and
many more leased third-party colocation data center spaces
spreading throughout the world \cite{Google_SustainabilityReport_2023}. By renting virtual machines on public clouds,
even a small business can flexibly choose its deployment region
and place its computing workloads accordingly.
As such, GLB is an important and common knob
that can spatially balance computing workloads' energy demand as well as environmental footprint across different locations.

As a concrete example, we consider
moving AI inference
workloads around from one data center to another and exploit
equity-aware GLB to mitigate AI's environmental inequity.
To achieve equitable distribution of AI's environmental cost, we consider the notion of \emph{minimax} fairness.
Mathematically, denoting $x_{i,t}$ as the
amount of AI workloads processed in data center $i$
at time $t$ and $E_{i,t}(x_{i,t})$ as the resulting regional environmental
cost (e.g., due to water consumption \cite{Shaolei_Water_AI_Thirsty_arXiv_2023_li2023making} and air/thermal/waste pollution
from non-renewable energy \cite{US_EPA_Electricity_LocalImpact}),
we consider an equity-aware objective:
$\sum_{t=1}^T\sum_{i} cost_{i,t}(x_{i,t}) + \lambda\cdot\max_{i}\left[\sum_{t=1}^TE_{i,t}(x_{i,t})\right]$, where the first term is the traditional GLB cost 
(e.g., total carbon/water footprint and energy cost) specified
based on the prior literature \cite{Shaolei_Water_SpatioTemporal_GLB_TCC_2018_7420641}, the second term ``$\max_{i}\left[\sum_{t=1}^TE_{i,t}(x_{i,t})\right]$'' 
 serves as the equity regularizer by reducing the highest regional
environmental cost, 
and $\lambda\geq0$ is the weight.

\begin{wrapfigure}[9]{r}{.7\textwidth}
	\centering
    \vspace{-4mm}
    \scriptsize
    \centering
    \begin{tabular}{c|c|c|c|c|c|c} 
    \toprule
    {\multirow{2}{*}{\textbf{GLB}}}& {\multirow{2}{*}{\textbf{Metric}}} & \multicolumn{5}{c}{\textbf{Algorithm}}                                                          \\ 
    \cline{3-7}
            &                        &  \costalg       &  \carbonalg            
    & \distancealg         & {\ouralgoffline} & \textbf{\ouralg}\\ 
    \hline
    \multirow{3}{*}{Full}            & \textbf{Cost} (US\$)                            & 29170 & 45535  & 47038 & {33669}& \textbf{33752}\\ 
    \cline{2-7}
    & {\textbf{PAR} (Water)}      &  1.71 & 1.85  &  1.44 & 1.27& \textbf{1.37} \\
    \cline{2-7}
    & {\textbf{PAR} (Carbon)}   & 1.68 & 1.70  & 1.41 & 1.13 & \textbf{1.22} \\  
    \hline
    \multirow{3}{*}{Partial}  & \textbf{Cost} (US\$)    & 29659 & 45535   & 47038 & {34186} &\textbf{34162} \\  
    \cline{2-7}
    & {\textbf{PAR} (Water)}  & 1.72 & 1.84 & 1.44 & {1.30} &\textbf{1.38} \\  
   \cline{2-7}
    & {\textbf{PAR} (Carbon)}   & 1.69 & 1.71 & 1.41 & 1.12 & \textbf{1.22}\\   \bottomrule
    \end{tabular}
    \vspace{-0.2cm}
    \caption{\textit{Comparison of GLB algorithms in terms
    of the total energy cost and the normalized water/carbon peak-to-average ratio (\emph{{PAR})}. Details in~\cite{Shaolei_Equity_GLB_Environmental_AI_eEnergy_2024}.}}  
    \label{table:preliminary_environmental_equity_AI}
 \end{wrapfigure}
\noindent\textit{A snapshot of results.} 
We run a 
simulation based on the BLOOM model (a large language model) inference trace deployed in 10 different data centers 
throughout the world
and show a snapshot of our results in Table~\ref{table:preliminary_environmental_equity_AI}. 
The details of the simulation are available in \cite{Shaolei_Equity_GLB_Environmental_AI_eEnergy_2024}. 
We consider both \emph{full} GLB (i.e., each request can
be flexibly routed to any data center)
and \emph{partial} GLB (i.e., each request can only
be routed to a subset of data centers depending on its originating
location). Compared
to common baseline algorithms that simply minimize the total energy cost
(\costalg),
carbon emission (\carbonalg) or workload-to-data center distance (\distancealg),
our algorithm (called \ouralgoffline)
can effectively mitigate the environmental inequity
by reducing the ratio of the maximum to the average regional environmental footprint. Importantly, while there is an inevitable
conflict between minimizing the total cost/environmental footprint
and addressing the environmental inequity, \ouralgoffline
can still keep the total cost reasonably low.
 Additionally,
we study a simple online algorithm (called \ouralg) based on dual mirror descent
 to show the potential
of mitigating environmental inequity in an online setting. 
While there is a gap between \ouralg and \ouralgoffline due
to online informational constraints,
\ouralg 
outperforms the equity-unaware baseline algorithms in terms of the
environmental footprint's peak-to-average ratio, demonstrating the potential of online GLB
to mitigate AI's environmental inequity.

\subsection{Challenges}

While equity-aware GLB can potentially 
mitigate AI's environmental inequity,
the equity regularizer ``$\max_{i}\left[\sum_{t=1}^TE_{i,t}(x_{i,t})\right]$''
fundamentally
separates our problem from the existing sustainable GLB approaches
and creates substantial algorithmic challenges.
Specifically, the equity cost
``$\max_{i}\left[\sum_{t=1}^TE_{i,t}(x_{i,t})\right]$''
is unknown until the end of $T$ time slots, but complete future information (e.g., future workload arrivals
and water efficiency) may not be perfectly known in advance. 
Moreover, even though prediction is often available in practice, it 
may not be accurate, and its untrusted nature means we cannot simply
take the prediction as if it were the ground truth.

Additionally, the traditional design of online competitive algorithms often
focuses on guaranteeing the worst-case performance robustness.
But, the resulting average performance can be far from optimal due to
the conservativeness needed to address potentially worst instances.
By contrast, machine learning (ML) based optimizers, e.g.,
reinforcement learning policies, can improve
the average performance of online decision-making by exploiting 
rich historical data and statistical information, but they typically sacrifice the strong performance robustness needed by real 
AI systems, especially when there is a distributional shift,
the ML model capacity is limited, and/or inputs are adversarial.
Thus, in order to achieve the best of both worlds while pursuing online equitable-aware GLB, we have to carefully balance the usage of traditional competitive algorithms and ML-based optimizers by designing new learning-augmented online
algorithms.

\section{Future Directions}

We discuss a few future directions to 
leverage system knobs
for environmentally equitable AI.

\textbf{Coordinated scheduling
 of AI training and inference.} While AI inference
offers spatial flexibility, AI model training has great \textit{temporal} scheduling flexibility as we can choose \emph{when} to train the AI models in
a stop-and-go manner. 
We can also choose where to perform AI model training and even possibly change the 
locations in the middle of the training process. Thus,  
a potential direction is to explore coordinated scheduling
of AI training and inference tasks to fairly distribute AI's overall
environmental costs across different regions.

\textbf{Joint optimization of IT and non-IT resources.}
Data centers have increasingly begun to install on-site carbon-free energy,
such as solar power, to partially power the workloads and lower the environmental footprint \cite{Google_CarbonAwareComputing_PowerSystems_2023_9770383}. 
However, renewables are often intermittent, and the available energy
storage capacity is finite.
Thus, how to optimize AI demand response 
given intermittent renewables is challenging, yet
worth investigating for addressing AI's environmental inequity.

\textbf{Holistic control of system knobs.} In addition to GLB,
 a rich set of system knobs are available and offer flexible tradeoffs, 
 such as dynamic model selection for inference, turning servers on/off, and resource allocation to different AI tasks. For example, 
different AI models can exhibit different energy-accuracy tradeoffs
for the same task. 
Holistic control of these system knobs holds enormous
potential to curb AI's resource usage and mitigate environmental
inequity, but also presents additional challenges due to the significantly
enlarged decision space.

\textbf{Theoretical foundations.}  
Optimizing a variety of system knobs
for environmentally equitable AI has its roots in \emph{fair} decision-making, which
is a classical area that bridges computer systems and algorithms and enjoys a long history with rich theoretical results \cite{mo2000fair,srikant2013communication} and prominent production deployments.
However, this classic literature primarily focuses on algorithms that ensure that different job or flow types receive a fair share of system resources, e.g., CPU, memory, etc. Tackling the challenges raised by environmental inequity in modern planet-scale AI systems requires a revisit to the algorithmic foundations
and the development of new theoretical tools, which can systematically capture the conflicts between traditional measures of performance, such as accuracy and latency, with measures of emerging importance, such as environmental equity.
Thus, it is crucial to build new theoretical foundations to support the design of environmentally equitable AI.  
\section{Conclusion}

In light of AI's wildly different environmental costs across different regions,
we advocate \emph{environmental equity} as a priority
for the management of future AI systems.
We present the potential opportunities and algorithmic challenges of tapping into
AI workloads' scheduling flexibility and leveraging equity-aware
GLB to mitigate AI's environmental inequity.
Finally, we discuss a few future directions to unleash
the full potential of system knobs
for environmentally equitable AI,
including coordinated scheduling
 of AI training and inference, joint optimization of IT and non-IT resources,
 holistic control of system knobs, and building theoretical foundations.
Our proposal of environmental equity pushes forward
 the boundaries of existing system management for sustainable AI
 and also adds a unique dimension
to AI fairness,
collaboratively building equitable and responsible AI.

\section*{Acknowledgement}
The work of Mohammad Hajiesmaili is supported by NSF CNS-2325956, CAREER-2045641, CPS-2136199, CNS-2102963, CNS-2106299, and NGSDI-2105494.
The work of Shaolei Ren is supported by NSF CCF-2324916.
The work of Ramesh K. Sitaraman is supported by NSF CNS-2325956, NGDSI-2105494, and CNS-1763617. The work of Adam Wierman is supported by NSF CCF-2326609, CNS-2146814, CPS-2136197, CNS-2106403, and NGSDI-2105648.

\bibliographystyle{ACM-Reference-Format}

\end{document}